\documentclass[10pt,twocolumn,letterpaper]{article}

\usepackage{arxiv}
\usepackage{times}
\usepackage{epsfig}
\usepackage{graphicx}
\usepackage{amsmath}
\usepackage{amssymb}

\usepackage{algorithm}
\usepackage{algpseudocode}
\algrenewcommand\algorithmicindent{.9em}%

\usepackage{comment}
\usepackage[british,UKenglish,USenglish,english,american]{babel}
\usepackage{multirow}
\usepackage{xcolor}
\usepackage{mathtools}

\newcommand{\tabincell}[2]{\begin{tabular}{@{}#1@{}}#2\end{tabular}}

\usepackage[pagebackref=true,breaklinks=true,letterpaper=true,colorlinks,bookmarks=false]{hyperref}

\begin{document}

\title{Integrated Object Detection and Tracking with Tracklet-Conditioned Detection}

\author{Zheng Zhang$^1$\thanks{Equal contribution. \dag This work is done when Dazhi Cheng and Xizhou Zhu are interns at Microsoft Research Asia.}\quad Dazhi Cheng$^{1,2*\dag}$ \quad Xizhou Zhu$^{1,3*\dag}$ \quad Stephen Lin$^1$ \quad Jifeng Dai$^1$ \vspace{8pt}\\
	$^1$Microsoft Research Asia\\
	$^2$Beijing Institute of Technology\\
    $^3$University of Science and Technology of China\\
	{\tt\small \{zhez,v-dachen,stevelin,jifdai\}@microsoft.com} \\
	{\tt\small ezra0408@mail.ustc.edu.cn} \\
}

\maketitle

\begin{abstract}

Accurate detection and tracking of objects is vital for effective video understanding. In previous work, the two tasks have been combined in a way that tracking is based heavily on detection, but the detection benefits marginally from the tracking. To increase synergy, we propose to more tightly integrate the tasks by conditioning the object detection in the current frame on tracklets computed in prior frames. With this approach, the object detection results not only have high detection responses, but also improved coherence with the existing tracklets. This greater coherence leads to estimated object trajectories that are smoother and more stable than the jittered paths obtained without tracklet-conditioned detection. Over extensive experiments, this approach is shown to achieve state-of-the-art performance in terms of both detection and tracking accuracy, as well as noticeable improvements in tracking stability. 

\end{abstract}

\section{Introduction}

Detection and tracking of moving objects is an essential element of many video understanding tasks, such as visual surveillance, autonomous navigation, and video captioning. Different from the more commonly addressed problem of object detection in still images, the additional temporal dimension in the video case introduces challenges that arise from scene dynamics. As an object moves, its appearance can vary due to occlusions, pose changes, and illumination differences. Imaging-related degradations such as motion blur and video defocus may affect object appearance as well. These factors collectively complicate the task of discovering objects and following their trajectories in a scene.

A common practice among existing methods for object detection and tracking is to detect objects in each frame independently and then link the detected objects across frames to form tracklets~\cite{xiang2015learning,kang2016object,chu2017online,feichtenhofer2017detect}. Applying detection and tracking in this sequential manner is of appealing simplicity. But unlike how detection assists tracking in this approach, there are no means for tracking to aid detection. Some methods attempt to address this issue by using tracklets to propagate detection bounding boxes from previous frames to the current frame, and then add these boxes to those produced by the detector~\cite{xiang2015learning,kang2016object,chu2017online}. However, with this {\em late integration} of tracking into the detection process, the tracking has no effect on the object detector itself. Rather, tracking exerts its influence only {\em after} the object detector has computed its bounding box results.

The disjoint design can be partially attributed to the relatively independent development of video object detection and multi-object tracking techniques. In the research of video object detection, the focus is on improving the per-frame object detection accuracy, while employing off-the-shelf trackers for post-processing~\cite{kang2016object, zhu2017flow}. Meanwhile, for research on multi-object tracking, the detection results are usually assumed to be given by external object detectors applied on individual frames~\cite{huang2008robust,zhang2008global,berclaz2011multiple}. Such decoupling simplifies research for each task, but misses the benefit of integrating detection and tracking.

\begin{figure*}[t]
\centering
\includegraphics[width=1.0\textwidth]{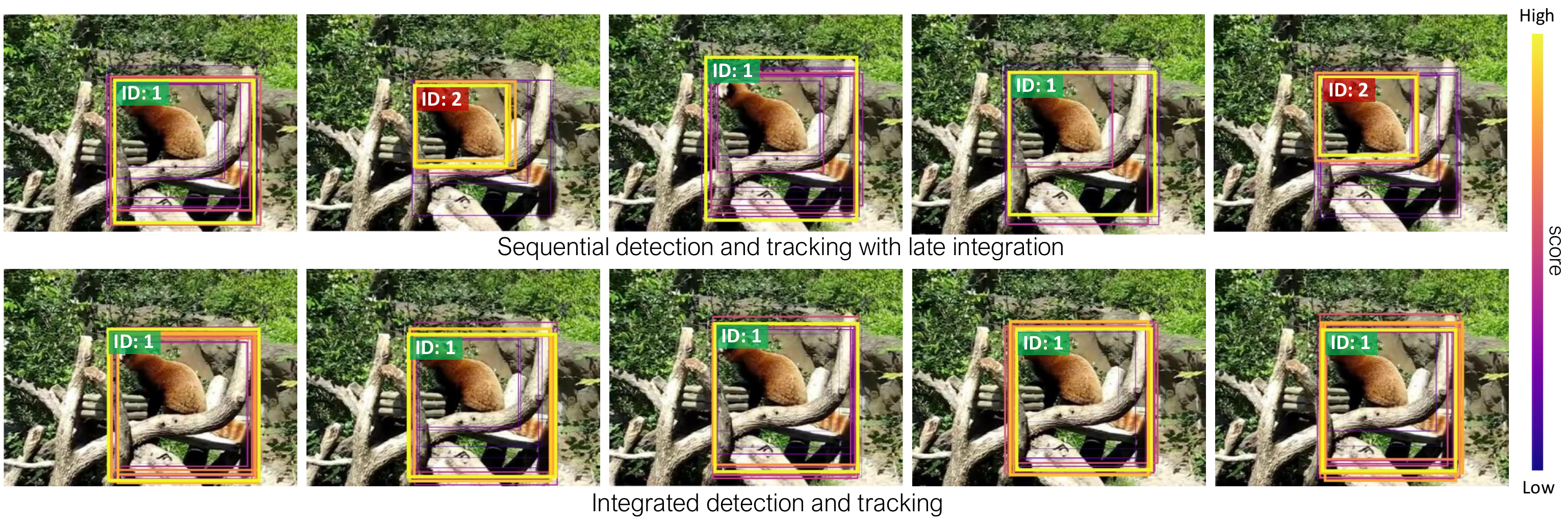}
\caption{Visualized detection and tracking results by the previous late integration (top row) and the proposed early integration (bottom row) approaches. Shown are scored bounding boxes prior to non-maximum suppression (NMS), where the boxes are colored according to the corresponding category scores. The highest scored bounding box at each image is kept after NMS as the detection result, and is associated to existing tracklets (or initiates a new tracklet if association fails). The tracklet ID is indicated at the top-left corner of the detection box. More accurate and stable results are generated by the proposed approach of integrating detection and tracking early.}
\label{fig:comparison_sequential_integrated}
\vspace{0em}
\end{figure*}

In this paper, we present an approach in which detection and tracking are more closely intertwined through an {\em early integration} of the two tasks. Instead of simply aggregating two sets of bounding boxes that are estimated separately by the detector and tracker, a single set of boxes is generated jointly by the two processes by conditioning the outputs of the object detector on the tracklets computed over the prior frames. In this way, the resulting detection boxes are both consistent with the tracklets and have high detection responses, instead of often having just one or the other in late integration techniques.

This advantage is illustrated in Fig.~\ref{fig:comparison_sequential_integrated}, which shows an example of detection boxes obtained with late integration as done in~\cite{kang2016object},
and with early integration via our tracklet-conditioned detection. Due in part to the aforementioned challenges of object detection in video, the boxes that have the highest detection scores without consideration of tracking may lie at various locations that deviate from the corresponding tracklet. 
Including boxes from late integration provides additional candidates, but they may not coincide closely with the actual object location due to errors in optical flow.
With our tracklet-conditioned detection, temporal cues compiled over multiple frames can robustly guide the detector in a way that can compensate for variabilities in the detection of moving objects.

\begin{algorithm}[t]
\caption{Online sequential detection and tracking.}
\begin{algorithmic} 
\State \textbf{input}: video frames $\{\mathbf{I}_t\}_{t=0}^T $
\State $\mathbf{B}_0 \coloneqq DetectOnImage(\mathbf{I}_0)$
\State initialize the tracklets $\mathbf{D}_0$ from $\mathbf{B}_0$
\For{$t=1$ \textbf{to} $T$}
\State $\mathbf{B}_t \coloneqq DetectOnImage(\mathbf{I}_t)$

\State $\mathbf{B}'_t \coloneqq PropagateBox(\mathbf{D}_{t-1})$ \quad Optional
\State $\mathbf{B}_t \coloneqq [\mathbf{B}_t, \mathbf{B}'_t]$ \quad Optional
\State $\mathbf{B}_t \coloneqq NMS(\mathbf{B}_t)$
\State $\mathbf{D}_t \coloneqq AssociateTracklet(\mathbf{D}_{t-1}, \mathbf{B}_t)$
\State $\mathbf{B}_t \coloneqq RescoreBox(\mathbf{D}_{t})$ \quad Optional
\EndFor
\State \textbf{output}: all tracklets $\mathbf{D}_T$ and all boxes $\{\mathbf{B}_t\}_{t=0}^T$
\end{algorithmic}
\label{alg.sequential_detect_track}
\end{algorithm}

A natural outcome of tracklet-conditioned detection is increased stability in tracking. Besides generating detection boxes that more closely adhere to the target object, the conditioning also results in smoother trajectories where the detection boxes overlap the moving object in a consistent manner, as shown in Fig.~\ref{fig:comparison_sequential_integrated} and detailed in Sec.~\ref{sec:discussion}. This property is beneficial for applications such as live compositing of virtual makeup on faces, where a lack of stability would produce unwanted jittering of the makeup relative to the face.

We show how tracklet-based conditioning can be applied within a modern two-stage detector, employing it in both region proposal generation and classification. Through comprehensive evaluation on the Image VID~\cite{ILSVRC15} and MOT~\cite{leal2015motchallenge} datasets, it is shown that this provides state-of-the-art performance on both object detection and tracking. Noticeable gains in tracking stability are achieved as well. The code for this technique will be released.

\section{Integrated Object Detection and Tracking}
\subsection{Background}

Given a video of multiple frames $\mathbf{I}_t, t=0, \ldots, T$, our goal is to detect and to track all the object instances within it up to time $t$, which we denote as $\mathbf{D}_t$. $\mathbf{D}_t = \{<d_j^t, c_j^t>\}, j=1, \ldots, m$, where $d_j^t$ denotes the $j$-th tracklet, and $c_j^t$ denotes its corresponding category. For a tracklet $d^t$, it is composed of a set of bounding boxes detected on individual frames up to time $t$, as $d^t = [b_k^{t_k}]$, where $b_k^{t_k}$ is the $k$-th bounding box in $d^t$ at frame $t_k$, where $t_k\le t$.

A scheme widely adopted in previous work~\cite{xiang2015learning,kang2016object,chu2017online,feichtenhofer2017detect} is sequential detection and tracking, outlined in Algorithm 1. Here we describe an online variant of the algorithm. Given a new video frame $\mathbf{I}_{t}$, an object detector for individual images is first applied to produce per-frame detection results $\mathbf{B}_t^{\text{}}\coloneqq DetectOnImage(\mathbf{I}_t)$, where $\mathbf{B}_t$ denotes a set of bounding boxes together with their corresponding category scores. Non-maximum suppression is then applied to remove redundant bounding boxes, resulting in $\mathbf{B}_t \coloneqq NMS(\mathbf{B}_t)$. Then the tracking algorithm associates the existing tracklets $\mathbf{D}_{t-1}$ to the detection results $\mathbf{B}_t$, producing tracklets up to frame $\mathbf{I}_t$ as $\mathbf{D}_t\coloneqq AssociateTracklet(\mathbf{D}_{t-1}, \mathbf{B}_t)$. Finally, the algorithm outputs all the tracklets $\mathbf{D}_T$ up to time $T$. 

To improve performance, two optional techniques are widely used as adds-on to better exploit tracklet information: (1) Box propagation, where detected boxes in the existing tracklets $\mathbf{D}_{t-1}$ are propagated to the current frame $\mathbf{I}_t$, usually with the aid of flow information, to get boxes $\mathbf{B}'_t \coloneqq PropagateBox(\mathbf{D}_{t-1})$. The propagated boxes are concatenated with the per-image detected boxes as $\mathbf{B}_t \coloneqq [\mathbf{B}_t, \mathbf{B}'_t]$. The concatenated boxes undergo non-maximum suppression and are associated to the existing tracklets. This technique can be helpful when bounding boxes are not reliably detected in the new frame $\mathbf{I}_t$. (2) Box rescoring, a post-processing step to obtain more accurate classification scores for the detected boxes. For a bounding box newly associated to a tracklet, its score is set to the average score of all the bounding boxes that compose the tracklet. Here we denote this operation as $\mathbf{B}_t \coloneqq RescoreBox(\mathbf{D}_t)$.

Including box propagation and/or box rescoring leads to better integration of detection and tracking. However, these techniques allow tracking to impact detection at only a late stage, after the per-image detection boxes are fixed. As a result, the detector cannot take full advantage of the tracking information.

\subsection{Tracklet-Conditioned Detection Formulation}
\label{sec:formulation}

We aim at improving per-frame detection results through early integration of object detection and tracking. Our goal is for detection to exploit not only the image appearance of the current frame, but also information from tracklets recovered in the previous frames. We refer to this approach as tracklet-conditioned detection.

The problem can be formulated as: Given a set of candidate boxes $\{b_{i}^{t}\}$\footnote{The candidate boxes can be either dense sliding windows / anchor boxes in the first stage of two-stage object detectors, or sparse region proposals in the second stage.} on frame $\mathbf{I}_t$, where $b_i^t$ specifies the 4-D coordinates of the $i$-th box, together with the tracklets $\{d_j^{t-1}\}_{j=1}^{m}$ up to frame $\mathbf{I}_{t-1}$, classify each box to different categories (including background) by estimating the score $P(c|b_{i}^{t},\{d_j^{t-1}\})$. Based on the intuition that a candidate box should more likely take labels consistent with tracklets it is more likely to be associated with, the score is further decomposed to be conditioned on each tracklet, as
\begin{equation}
P(c|b_{i}^{t},\{d_j^{t-1}\}) = \sum_{j=0}^{m}w(b_i^t,d_j^{t-1})P(c|b_i^t,d_j^{t-1}),
\label{eq:tracklet_conditioned_detection}
\end{equation}
where $w(b_i^t,d_j^{t-1})$ specifies the association weight between box $b_i^t$ and tracklet $d_j^{t-1}$. To account for newly detected objects that do not appear in existing tracklets, we include a null tracklet $d_0^{t-1}$, as detailed at the end of this subsection.

The score $P(c|b_i^t,d_j^{t-1})$ is estimated based on both the appearance of the current frame and information from previous tracklets, as
\begin{equation}
P(c|b_i^t,d_j^{t-1}) \propto \exp(\log P_{\textbf{det}}(c|b_i^t)+\alpha\log P_{\textbf{tr}}(c|d_j^{t-1})),
\end{equation}
where $P_{\textbf{det}}(c|b_i^t)$ is predicted by the per-image object detector on $\mathbf{I}_t$, $P_{\textbf{tr}}(c|d_j^{t-1})$ is the classification probability for tracklet $d_j^{t-1}$, and the hyper-parameter $\alpha$ balances the two log-likelihood terms ($\alpha=1$ by default). $P(c|b_i^t,d_j^{t-1})$ is normalized over all the categories, by $\sum_{c=0}^{C} P(c|b_i^t,d_j^{t-1}) = 1$, where $C$ denotes the number of foreground object categories, plus one for the background ($c=0$). $P_{\textbf{tr}}(c|d_j^{t-1})$ is defined on the classification scores of all the bounding boxes assigned to tracklet $d_j^{t-1}$, in a running average fashion. Suppose tracklet $d_j^{t}$ ($j>0$) is composed of box $b_{k}^{t}$ and tracklet $d_j^{t-1}$, then $P_{\textbf{tr}}(c|d_j^t)$ is computed as
\begin{equation}
P_{\textbf{tr}}(c|d_j^{t}) = \frac{P(c|b_{k}^{t},\{d_j^{t-1}\}) + \beta P_{\textbf{tr}}(c|d_j^{t-1}) \text{len}(d_j^{t-1})}{1+\beta \text{len}(d_j^{t-1})}
\label{eq:tracklet_rescore}
\end{equation}
where $\beta$ is an exponential decay parameter ($\beta = 0.99$ by default), and $\text{len}(d_j^{t-1})$ denotes the trajectory length of $d_j^{t-1}$. 

The association weight $w(b_i^t,d_j^{t-1})$ is defined based on the intuition that box $b_i^t$ is more likely to be associated to a tracklet that is visually similar: 
\begin{equation}
w(b_i^t,d_j^{t-1}) \coloneqq \exp(\gamma\cos(\mathcal{E}(b_i^t), \mathcal{E}(d_j^{t-1}))) \quad j>0,
\label{eq:association_weight_exist}
\end{equation}
where $\mathcal{E}(b_i^t)$ and $\mathcal{E}(d_j^{t-1})$ are embedding features (128-D in our work) that encode the visual appearance of box $b_i^t$ and tracklet $d_j^{t-1}$ respectively, which are generated as described in Section~\ref{sec:two_stage}. The cosine similarity between the embedding features is calculated and modulated by hyper-parameter $\gamma$ (set to $8$ in this paper) to be the log-likelihood of the association weight.

It is worth noting that new objects may appear in a video frame, and these objects will not be associated with any existing tracklets. To handle these cases, a null tracklet $d_0^{t-1}$ is introduced. For every candidate box, its association weight with $d_0^{t-1}$ is set to a constant, as
\begin{equation}
w(b_i^t,d_0^{t-1}) \coloneqq \exp(R),
\label{eq:association_weight_null}
\end{equation}
where $R=0.3$ in this paper. The association weights defined in Eq.~\eqref{eq:association_weight_exist} and Eq.~\eqref{eq:association_weight_null} are further normalized over all the tracklets, as
\begin{equation}
w(b_i^t,d_j^{t-1}) \coloneqq \frac{w(b_i^t,d_j^{t-1})}{\sum_{k=0}^{m}w(b_i^t,d_k^{t-1})}.
\label{eq:association_weight_tracklet}
\end{equation}
Thus, when a candidate box has low association weights with all the existing tracklets, its normalized association weight with the null tracklet will be high. For the null tracklet, its classification probability is set to a uniform distribution over all the categories, as
\begin{equation}
P_{\textbf{tr}}(c|d_0^{t}) = \frac{1}{C+1}.
\end{equation}

\subsection{Tracklet-Conditioned Two-stage Detectors}
\label{sec:two_stage}

\begin{figure*}[t]
\centering
\includegraphics[width=0.98\textwidth]{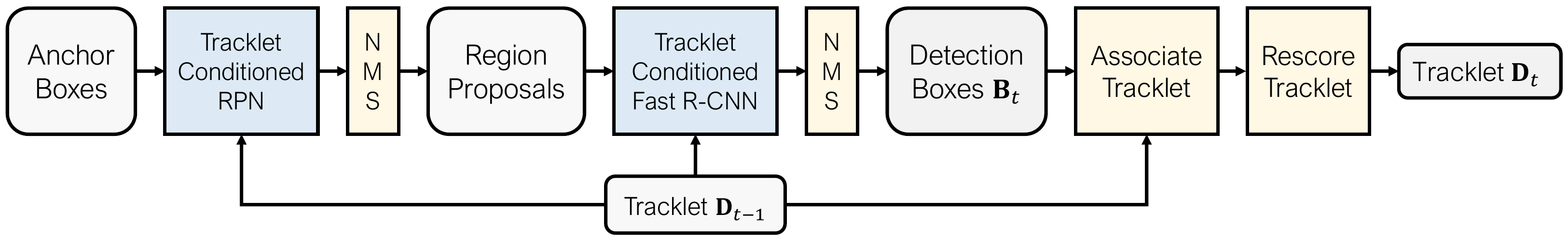}
\caption{Tracklet-conditioned two-stage detectors.}
\label{fig:tracklet_two_stage}
\vspace{-1em}
\end{figure*}

The proposed tracklet-conditioned detection algorithm can be readily applied in state-of-the-art object detectors. In this paper, we incorporate it into the two-stage Faster R-CNN~\cite{ren2015faster} + ResNet-101~\cite{he2016deep} detector, with OHEM~\cite{shrivastava2016training}. For this baseline, following the practice in~\cite{dai2017deformable}, all the convolutional layers in ResNet-101 are applied on the whole input image. The effective stride in the conv5 blocks is reduced from 32 to 16 pixels to increase feature map resolution. The RPN~\cite{ren2015faster} head is added on top of the conv4 features of ResNet-101. The Fast R-CNN~\cite{girshick2015fast} head is added on top of the conv5 features, and is composed of RoIpooling and two fully-connected (fc) layers of 1024-D, followed by the classification and the bounding box regression branches.

The tracklet-conditioned two-stage detector is exhibited in Figure~\ref{fig:tracklet_two_stage}. The tracklet conditioning in the second stage is relatively straightforward, with the equations in Section~\ref{sec:formulation} applied on sparse region proposals. $P_{\textbf{det}}(c|b_i^t)$ is predicted by the classification branch of the Fast R-CNN detection head. The box embedding $\mathcal{E}_{\text{s2}}(b_i^t)$ (of the second stage) is computed by attaching a branch (consisting of a fully-connected layer) to the Fast R-CNN head, sibling to the classification and bounding box regression branches. The tracklet embedding $\mathcal{E}_{\text{s2}}(d_j^{t})$ ($j>0$) is updated based on the embedding features of the boxes associated to it, as
\begin{equation}
\mathcal{E}_{\text{s2}}(d_j^{t}) = 
\begin{cases}
\eta \mathcal{E}_{\text{s2}}(b_k^t) + (1-\eta) \mathcal{E}_{\text{s2}}(d_j^{t-1}) \quad & \text{if } t>0, \\
\mathcal{E}_{\text{s2}}(b_k^0) \quad & \text{otherwise,}
\end{cases}
\label{eq:tracklet_embedding}
\end{equation}
where $b_k^t$ denotes the detection box associated to tracklet $d_j^{t}$ at time $t$, and $\eta$ is the update weight parameter ($\eta = 0.8$ by default). The box embedding features $\mathcal{E}_{\text{s2}}(b_i^t)$ are compared to the tracklet embedding features $\mathcal{E}_{\text{s2}}(d_j^{t-1})$ by Eq.~\eqref{eq:association_weight_exist} to obtain the association weights for the second stage. 

We further apply the tracklet-conditioned detection in the first stage, to make use of tracklet information for improving region proposal quality. Compared to the application in the second stage, the key differences are that the candidate boxes are dense anchor boxes, and only two categories are involved, namely foreground and background. Given an anchor box $b_i^t$, its foreground probability $P(\text{fg}|b_{i}^{t},\{d_j^{t-1}\})$ is estimated by Eq.~\eqref{eq:tracklet_conditioned_detection}, with $P_{\textbf{det}}(\text{fg}|b_i^t)$ predicted by the RPN classification branch and $P_{\textbf{tr}}(\text{fg}|d_j^{t-1})$ computed as
\begin{equation}
P_{\textbf{tr}}(\text{fg}|d_j^{t-1}) = \sum_{c=1}^{C}P_{\textbf{tr}}(c|d_j^{t-1}),
\end{equation}
which is the summation of the probability $P_{\textbf{tr}}(c|d_j^{t-1})$ over all the foreground categories ($c>0$).

To derive the association weights for the first stage, two additional branches are added for producing the embedding features. The embedding features $\mathcal{E}_{\text{anchor}}(b_i^t)$ for the dense anchor boxes are computed following the design in \cite{li2018high}, via a sibling branch (consisting of a $1\times1$ convolution) added to the RPN classification branch. Supposing there are $K$ anchors at each location and the embedding features are 128-D, the output of the embedding branch is of dimension $128\times K$. The tracklet embedding features $\mathcal{E}_{\text{s1}}(d_j^{t})$ of the first stage are computed in a manner similar to those of the second stage. An additional branch is added to the Fast R-CNN head to produce $\mathcal{E}_{\text{s1}}(b_i^t)$. After the RoIpooling layer, two additional fc layers of 1024-D are added (sibling to the existing two fc layers), followed by one fc layer to produce $\mathcal{E}_{\text{s1}}(b_i^t)$. Here, we tried different network designs for producing $\mathcal{E}_{\text{s1}}(b_i^t)$, as shown in Table~\ref{table.parameter_abalation}. We find that adding two fc layers to reduce correlation between the embedding features of the two stages is beneficial for accuracy. Given $\mathcal{E}_{\text{s1}}(b_i^t)$, $\mathcal{E}_{\text{s1}}(d_j^{t})$ is obtained by applying Eq.~\eqref{eq:tracklet_embedding} (replacing the subscript s2 by s1 in the equation). Finally, the anchor box embedding features $\mathcal{E}_{\text{anchor}}(b_i^t)$ are compared to the tracklet embedding features $\mathcal{E}_{\text{s1}}(d_j^{t-1})$ by Eq.~\eqref{eq:association_weight_exist} to obtain the association weights for the first stage.

\subsection{Training and Inference}

\begin{algorithm}[t]
\caption{Online integrated detection and tracking.}
\begin{algorithmic} 
\State \textbf{input}: video frames $\{\mathbf{I}_t\}_{t=0}^T$
\State $\mathbf{B}_0 \coloneqq DetectOnImage(\mathbf{I}_0)$
\State initialize the tracklets $\mathbf{D}_0$ from $\mathbf{B}_0$
\For{$t=1$ \textbf{to} $T$}

\State $\mathbf{B}_t \coloneqq TrackletCondDetect(\mathbf{I}_t, \mathbf{D}_{t-1})$
\State $\mathbf{B}_t \coloneqq NMS(\mathbf{B}_t)$
\State $\mathbf{D}_t \coloneqq AssociateTracklet(\mathbf{D}_{t-1}, \mathbf{B}_t)$
\State $\mathbf{D}_t \coloneqq RescoreTracklet(\mathbf{D}_{t})$
\EndFor
\State \textbf{output}: all tracklets $\mathbf{D}_T$ and all boxes $\{\mathbf{B}_t\}_{t=0}^{T}$
\end{algorithmic}
\label{alg.integrated_detect_track}
\end{algorithm}

\noindent\textbf{Inference}. Algorithm~\ref{alg.integrated_detect_track} presents the inference procedure for our integrated object detection and tracking with tracklet-conditioned detection. Given the input video frames $\{\mathbf{I}_t\}_{t=0}^T$, the per-image object detector is applied on the first frame $\mathbf{I}_0$ to produce detection boxes, $\mathbf{B}_0 \coloneqq DetectOnImage(\mathbf{I}_0)$. With these boxes, the tracklets $\mathbf{D}_0$ are initialized (one tracklet per box). Then for each subsequent frame $\mathbf{I}_t$, tracklet-conditioned detection is applied and followed by non-maximum suppression, as $\mathbf{B}_t \coloneqq TrackletCondDetect(\mathbf{I}_t, \mathbf{D}_{t-1})$ and $\mathbf{B}_t \coloneqq NMS(\mathbf{B}_t)$. 
As done in Algorithm~\ref{alg.sequential_detect_track}, the detected bounding boxes $\mathbf{B}_t$ are associated with the existing tracklets $\mathbf{D}_{t-1}$ by $\mathbf{D}_t \coloneqq AssociateTracklet(\mathbf{D}_{t-1}, \mathbf{B}_t)$. Then the obtained tracklets $\mathbf{D}_t$ are rescored as $\mathbf{D}_t \coloneqq RescoreTracklet(\mathbf{D}_{t})$, by applying Eq.~\ref{eq:tracklet_rescore}. Finally, the algorithm outputs all the tracklets $\mathbf{D}_T$ and all the boxes $\{\mathbf{B}_t\}_{t=0}^{T}$.

\vspace{1em}
\noindent\textbf{Training}. The network is trained to better detect objects based on image content and to better associate them across frames. Due to memory constraints, the forward pass in SGD training cannot be kept identical to that in inference. In each mini-batch, two consecutive frames from the same video, $\mathbf{I}_{t-1}$ and $\mathbf{I}_{t}$, are randomly sampled. In the forward pass, bounding boxes are detected on $\mathbf{I}_{t-1}$ based on image content only, as $\mathbf{B}_{t-1} \coloneqq DetectOnImage(\mathbf{I}_{t-1})$. The detected boxes are matched to the ground-truth annotations $\mathbf{B}_{t-1}^{\text{gt}}$ and $\mathbf{B}_{t}^{\text{gt}}$, on $\mathbf{I}_{t-1}$ and $\mathbf{I}_{t}$ respectively, to inject object detection loss and tracking loss.

The object detection loss is defined on $\mathbf{B}_{t-1}$ and $\mathbf{B}_{t-1}^{\text{gt}}$ in the same way as in conventional two-stage object detectors~\cite{ren2015faster,dai2016r}. It is composed of a foreground / background Softmax cross-entropy loss for region proposal scoring, an L1 regression loss for regressing proposal boxes, a $(C+1)$-way Softmax cross-entropy loss for detection scoring, and an L1 regression loss for regressing detected boxes.

The tracking loss is defined on $\mathbf{B}_{t-1}$ and the $\mathbf{B}_{t}^{\text{gt}}$ associated to $\mathbf{B}_{t-1}^{\text{gt}}$. For a detected box $b_{t-1} \in \mathbf{B}_{t-1}$, it is assigned to the ground-truth box $b^{\text{gt}}_{t-1} \in \mathbf{B}_{t-1}^{\text{gt}}$ having the highest IoU overlap with it. Let $b^{\text{gt}}_{t}$ be the ground-truth bounding box on the next frame that corresponds to the same object as $b^{\text{gt}}_{t-1}$. The tracking loss on $b_{t-1}$ is then defined as
\begin{equation}
\begin{aligned}
L &_{\text{track\_box}}(b_{t-1}, b^{\text{gt}}_{t-1}, b^{\text{gt}}_{t}) = \\
& \begin{cases}
(1 - \cos(\mathcal{E}(b_{t-1}), \mathcal{E}(b^{\text{gt}}_{t})))^2 \quad \text{if IoU}(b_{t-1}, b^{\text{gt}}_{t-1}) \ge 0.5\\
\max (0, \cos(\mathcal{E}(b_{t-1}), \mathcal{E}(b^{\text{gt}}_{t})))^2 \quad \text{otherwise} \\
\end{cases}
\end{aligned}
\end{equation}
which encourages the cosine similarity $\cos(\mathcal{E}(b_{t-1}), \mathcal{E}(b^{\text{gt}}_{t}))$ to be close to 1 if $b_{t-1}$ captures the same object as $b^{\text{gt}}_{t}$ ($\text{IoU}(b_{t-1}, b^{\text{gt}}_{t-1}) \ge 0.5$), and to be no more than 0 otherwise. The overall tracking loss is the summation of the loss values on all the detected boxes.

\subsection{Discussion}
\label{sec:discussion}

\noindent\textbf{Accuracy and robustness}
In the proposed approach, detection is enhanced by accounting for temporal information when determining the classification probabilities of the bounding boxes. In previous techniques, these probabilities $P_{\textbf{det}}(c|b_i^t)$ are obtained from the per-image object detector based solely on the appearance of frame $\mathbf{I}_t$. Object appearance variations and visual degradations in $\mathbf{I}_t$ can lead to significant distortions in the predicted probabilities of its detection boxes $\mathbf{B}_t$. To counteract these complications, our method takes advantage of the tracklets $\mathbf{D}_{t-1}$ from the previous frame, which model the visual appearance $\mathcal{E}(d_j^{t-1})$ and classification probabilities $P_{\textbf{tr}}(c|d_j^{t-1})$ of each object. As these tracklet attributes are computed over the full existing trajectory of an object, they provide a representation that is relatively robust to the appearance changes that may occur during object motion, while placing greater weight on more recent frames. The classification probabilities of a bounding box are influenced by the tracklets most similar to it, as determined from association weights. By taking advantage of tracklet information in this way, the classification scores of bounding boxes are more robustly obtained, leading to more accurate final boxes.

By comparison, late integration techniques typically incorporate temporal information by adding bounding boxes propagated from preceding frames by optical flow. These boxes are aggregated with the detector's bounding boxes just prior to NMS. As illustrated in Fig.~\ref{fig:comparison_sequential_integrated}, this is less ideal because the resulting boxes either have distorted classification probabilities (boxes from the detector) or rely on optical flow (boxes from tracklets) which can be inaccurate especially over the course of multiple frames or when there is background movement. Furthermore, since propagated boxes inherit the high classification scores of their corresponding tracklets, they may suppress more accurate boxes from the detector in the NMS. The difference in performance is examined in Sec.~\ref{sec:ablation}.

\vspace{0.5em}
\noindent\textbf{Stability}
Another key advantage of the proposed approach is the improvement of box localization stability across frames, as illustrated in Figure~\ref{fig:comparison_sequential_integrated}. Unstable localization is a commonly observed problem in video object detection and tracking, and such instability can be attributed to appearance change in different frames. For example, in Figure~\ref{fig:comparison_sequential_integrated}, suppose $b_i^{t-1}$ and $b_k^{t-1}$ are two candidate boxes properly covering the object `squirrel' on frame $\mathbf{I}_{t-1}$, but with slight shifts  respectively. For both boxes, the corresponding per-frame recognition scores $P_{\textbf{det}}(c|b_i^{t-1})$ and $P_{\textbf{det}}(c|b_k^{t-1})$ are high. But after NMS, only one of $b_i^{t-1}$ and $b_k^{t-1}$ would be kept. Suppose $P_{\textbf{det}}(c|b_k^{t-1})$ is slightly higher, so box $b_k^{t-1}$ is kept and associated with the existing tracklet to form $d_{j}^{t-1}$. On frame $\mathbf{I}_{t}$, suppose $b_{i'}^{t}$ and $b_{k'}^{t}$ are the highest overlapped candidate boxes with $b_i^{t-1}$ and $b_k^{t-1}$, respectively. Due to slight appearance changes on frame $\mathbf{I}_{t}$, $P_{\textbf{det}}(c|b_{i'}^{t})$ is slightly higher than $P_{\textbf{det}}(c|b_{k'}^{t})$, and thus $b_{i'}^{t}$ is kept after NMS. As a result, there is jitter between box $b_k^{t-1}$ and box $b_{i'}^{t}$ because of the sudden shift in box position relative to the actual object. If the jitter is large enough, the embedding features $\mathcal{E}(b_{i'}^t)$ can be quite different from those of the tracklet $\mathcal{E}(d_{j}^{t-1})$ that bounding box $b_k^{t-1}$ is associated with. Thus a mismatch occurs even though frames $\mathbf{I}_{t-1}$ and $\mathbf{I}_{t}$ are of high visual quality. Although box $b_{k'}^{t}$ could have been well associated with tracklet $d_{j}^{t-1}$ to generate a stable trajectory, it was already suppressed by NMS, thus becoming unavailable for consideration in tracklet association. 

Tracklet-conditioned detection can effectively remedy this issue. The candidate boxes on a new frame would be scored not only based on the per-frame appearance, but also based on their association weights with the existing tracklets. In the example of Figure~\ref{fig:comparison_sequential_integrated}, when scoring candidate boxes $b_{i'}^{t}$ and $b_{k'}^{t}$, the association weight $w(b_{k'}^t,d_j^{t-1})$ is high, and tracklet $d_j^{t-1}$ would cast a large vote on box $b_{k'}^{t}$. Thus the tracklet-conditioned score $P(c|b_{k'}^{t},\{d_j^{t-1}\})$ would be higher than those of the other boxes, and box $b_{k'}^{t}$ would be kept after NMS, generating a stable trajectory.

\subsection{Implementation Details}

In the procedure $AssociateTracklet$, we employ a modified version of maximum bipartite graph matching, an algorithm widely used in multi-object tracking systems~\cite{bae2014robust,milan2015joint,yu2016poi}. Given tracklets $\mathbf{D}_{t-1}$ and bounding boxes $\mathbf{B}_{t}$, a bipartite graph is generated in which nodes corresponding to $d_j^{t-1} \in \mathbf{D}_{t-1}$ and $b_i^{t} \in \mathbf{B}_{t}$ are on the two sides of the graph, respectively. An edge is added between $d_j^{t-1}$ and $b_i^{t}$ if there is overlap between the last box in $d_j^{t-1}$ and $b_i^{t}$, with their connection weight set to $\cos(\mathcal{E}(b_i^t), \mathcal{E}(d_j^{t-1}))$. There are no edges between non-overlapping tracklets and bounding boxes, so they will not be associated. To account for newly detected boxes which are not associated with any existing tracklet, a pseudo tracklet $d_i^{\text{pseudo}}$ is initialized for each such bounding box $b_i^{t}$. The nodes of the pseudo tracklets are added on the side of the existing tracklets in the bipartite graph. An edge is added between pseudo tracklet $d_i^{\text{pseudo}}$ and its corresponding bounding box $b_i^{t}$, with the connection weight set to 0. If the cosine similarity values between $b_i^{t}$ and existing tracklets are all low (less than 0), $b_i^{t}$ is likely to be associated with $d_i^{\text{pseudo}}$ and a new tracklet is formed. Finally, the standard Hungarian maximum matching algorithm~\cite{kuhn1955hungarian} is applied on the constructed bipartite graph to associate the bounding boxes to the tracklets.

For the procedure $PropagateBox$, we follow the implementation in~\cite{kang2016object}, but replace the OpenCV flow estimator~\cite{bradski2000opencv,lucas1981iterative} with the more recent FlowNet\_v2~\cite{ilg2017flownet} for more accurate correspondence estimation between frames.

\section{Related Work}

\paragraph{Object Detection in Images} Current leading object detectors are built on deep Convolutional Neural Networks (CNNs). They can be mainly divided into two families, namely, region-based two-stage detectors (\eg, R-CNN~\cite{girshick2014rich}, Fast(er) R-CNN~\cite{girshick2015fast,ren2015faster}, and R-FCN~\cite{dai2016r}) and one-stage detectors that directly predict boxes (\eg, YOLO~\cite{redmon2016you}, SSD~\cite{liu2016ssd}, and CornerNets~\cite{law2018cornernet}). 

We build our approach  on Faster R-CNN with ResNet-101 and OHEM, which is a state-of-the-art object detector.

\vspace{0.5em}
\noindent\textbf{Multiple Object Tracking (MOT)} Research on MOT primarily follow the ``sequential detection and tracking" paradigm, often under the setting where detection results are given by an external object detector and the focus is on correctly associating the detection boxes across frames. Various approaches to the association problem have been proposed, including but not limited to min-cost flow~\cite{zhang2008global, lenz2015followme}, energy models~\cite{milan2014continuous}, Markov decision processes~\cite{xiang2015learning}, node labeling~\cite{levinkov2017joint}, graph matching~\cite{bae2014robust}, and Graph Cut~\cite{zamir2012gmcp,tang2015subgraph}. In addition, recent works~\cite{sadeghian2017tracking,son2017multi,milan2017online,tang2017multiple,zhu2018online,kim2018multi,chu2017online,schulter2017deep} explore utilizing deep networks to better solve the association problem. In~\cite{leibe2007coupled,yu2008integrated,wu2012coupling,barbu2012simultaneous}, the authors seek to refine the detection and tracking results by various optimization formulations. Improved accuracies are reported, but the refinement is performed at a late stage on the results produced by off-the-shelf object detectors and trackers. 

In contrast to the previous research on multi-object tracking, we advocate a new paradigm of ``integrated object detection and tracking'', which aims to improve detection by considering tracking information and in turn further enhance tracking performance. The integration is at an early stage within the object detector. In this paper, tracking is performed simply by maximum bipartite graph matching, and we note that advances in MOT can benefit our method.

\vspace{0.5em}
\noindent\textbf{Single Object Tracking} In this classic vision problem, a single object is annotated in the first frame of an input video and the tracking algorithm aims to follow the specified object throughout subsequent frames. The major challenge lies in distinguishing the object from background clutter and occluding objects. To address these issues, recent works~\cite{zhang2017deep,li2018high,valmadre2017end,ma2015hierarchical,qi2016hedged,bertinetto2016fully, held2016learning, danelljan2017eco} leverage the strong representation power of deep networks, via either Siamese networks~\cite{bertinetto2016fully, li2018high, held2016learning} or correlation filters~\cite{valmadre2017end, DanelljanECCV2016, danelljan2017eco} based on network features. Such trackers are usually employed in MOT to provide the raw association weights of possible tracklet-box pairs. In this paper, we utilize a simple Siamese-network-based tracker to obtain the association weights, while also noting the benefits our method can reap from improvements in single object tracking.

\vspace{0.5em}
\noindent\textbf{Video Object Detection} Research on video object detection gained renewed interest with the introduction of the ImageNet VID benchmark~\cite{ILSVRC15}, which evaluates detection performance on individual frames. Numerous algorithms~\cite{kang2016object, zeng2018crafting, lee2016multi, zhu2017flow,feichtenhofer2017detect} and systems~\cite{yang2016vid,wei2017vid,deng2017vid} have been developed on it, with the main focus of improving per-frame object detection results by exploiting temporal information. In large, these works can be classified into box-level methods and feature-level techniques. 

Box-level methods~\cite{kang2016object, zeng2018crafting,feichtenhofer2017detect} primarily follow the ``sequential detection and tracking" approach. Bounding boxes are detected based on features from individual frames, and then are associated and rescored across frames. Prior to~\cite{feichtenhofer2017detect}, box-level techniques associate boxes across frames by employing external tracking modules. In~\cite{feichtenhofer2017detect}, for the first time, object detection and tracking modules share backbone features and are trained end-to-end. The network architecture design in our work follows~\cite{feichtenhofer2017detect} in sharing features. However, our inference procedure diverges from~\cite{feichtenhofer2017detect}, whose tracking module associates the detected boxes on individual frames at a late stage, like other ``sequential detection and tracking" techniques.

Feature-level techniques~\cite{zhu2017deep,zhu2017flow,zhu2018towards} enhance the quality of per-frame features by integrating temporal information, via flow-guided feature propagation from previous frames. This early exploitation of temporal cues leads to improved detection accuracy. We found that this technique can work in tandem with ours, by utilizing it to obtain the per-image detection scores in our method. Our experiments demonstrate the complementarity of these two approaches.

\section{Experiments}

\subsection{Experimental Settings}

We evaluate our method on two popular datasets. The first is ImageNet VID~\cite{ILSVRC15}, a large-scale video set where the object instances are fully annotated. Following the protocol in~\cite{lee2016multi, kang2016object}, we train our model on the union of the ImageNet VID and ImageNet DET training sets, and test our method on the ImageNet VID validation set. Evaluation is based on the ImageNet VID competition metrics. For detection, it is the mean average precision (mAP$^\text{det}$) score under a box-level IoU threshold of 0.5. For tracking, it is the mAP$^\text{track}$ score, under a box-level IoU threshold of 0.5 and temporal-level thresholds of [0.25, 0.5, 0.75]. All the ablations in this paper are performed on ImageNet VID.

The other dataset is 2D MOT 2015~\cite{leal2015motchallenge}, consisting of 11 training videos and 11 test videos with fully annotated object instances. On this benchmark, submission entries are split into public / private tracks, depending on whether they use the provided set of detection boxes or their own detector. As our work proposes a new detector, we compare it to entries in the private track. Due to the limited training samples, a common practice is to finetune the model on MOT train after training on large-scale external datasets~\cite{sadeghian2017tracking,son2017multi,bae2018confidence}. We note that since our approach integrates detection and tracking, we cannot train our detector on datasets consisting of cropped image patches (for person re-ID) as done in some sequential detection and tracking methods~\cite{sadeghian2017tracking, son2017multi,bae2018confidence}. So instead, we train our network on the COCO~\cite{lin2014microsoft} training and validation sets for object detection only, and finetune the whole network on the MOT training set for integrated detection and tracking. The standard evaluation metric of this dataset is MOTA, which combines false positives (FP) and false negatives (FN) for object detection, together with ID switch (IDSw) for tracking.

The hyper-parameters in training and inference on both datasets are presented in the Appendix. 

\subsection{Ablation Study}
\label{sec:ablation}

\setlength{\tabcolsep}{1pt}
\renewcommand{\arraystretch}{1.0}
\begin{table}[t]
\small
\begin{center}
\begin{tabular}{c|c|ccccc}
\hline
\multicolumn{2}{c|}{method} & \tabincell{c}{\scriptsize mAP$^{\text{det}}$} & \scriptsize mAP$^{\text{track}}$ & \scriptsize mAP$^{\text{track}}_{\text{slow}}$ & \scriptsize mAP$^{\text{track}}_{\text{med}}$ & \scriptsize mAP$^{\text{track}}_{\text {fast}}$\\
\hline
\multicolumn{2}{c|}{sequential baseline}    & 74.6& 65.2 & 79.3  & 54.8  & 33.5   \\
\hline
\multirow{2}{*}{\tabincell{c}{with late \\ integration}}
& +$PropagateBox$  & 75.2 & 65.9 &  80.3  & 53.3  & 38.9   \\
& ++$RescoreBox$   & 75.8 &65.9 & 80.3  & 53.3  & 38.9   \\
\hline
\multirow{3}{*}{integrated} & first stage only  & 75.4 &67.1  & 79.9  & 55.2  & 37.9   \\    
& second stage only     & 76.8 & 66.8 & 80.4  & 58.0 & 36.9  \\

& both stages             & \textbf{78.1} & \textbf{67.9} & \textbf{80.9}  & \textbf{58.1}  & \textbf{41.9}  \\
\hline
\end{tabular}
\end{center}
\caption{Ablation of key components of our integrated detection and tracking, and of sequential detection and tracking with late integration, on the ImageNet VID validation set.}
\vspace{-1.5em}
\label{table.component_abalation}
\end{table}

\setlength{\tabcolsep}{2pt}
\renewcommand{\arraystretch}{1.1}
\begin{table*}[t]
\small
\begin{center}
\begin{tabular}{c|ccc|ccc|ccc|ccc|ccc|ccc}
\hline
\multirow{3}{*}{design}& \multicolumn{15}{|c|}{parameters (default marked by *)} &\multicolumn{3}{|c}{box embedding features}\\
\cline{2-19}
& \multicolumn{3}{|c|}{$\alpha$} & \multicolumn{3}{|c|}{$\beta$} & \multicolumn{3}{|c|}{$\eta$} & \multicolumn{3}{|c|}{$\gamma$} & \multicolumn{3}{|c|}{$R$} & \multirow{2}{*}{\tabincell{c}{fully \\ shared}} & \multirow{2}{*}{\tabincell{c}{shared \\ 2fc}} & \multirow{2}{*}{\tabincell{c}{separate \\ 2fc}}  \\
\cline{2-16}
& 0.5 & 1.0* & 2.0 & 0.95 & 0.99* & 1.0 &0.7 &0.8* &0.9 & 4 & 8* & 16 &0.2 & 0.3* &0.4 & & & \\ 
\hline
mAP$^{\text{det}}$ & 78.4 & 78.1 & 76.7 &78.1 & 78.1 &78.1 & 78.0 & 78.1 & 78.1 & 76.9 & 78.1 &78.3 &78.1 &78.1 &78.1 & 77.2 &77.4 & 78.1 \\
\hline
mAP$^{\text{track}}$ & 66.7 & 67.9 &\ 67.9 &\ 67.5 &\ 67.9 &\ 67.4 & 66.8 &\ 67.9 &\ 67.4 &\ 67.5 &\ 67.9 &\ 67.8 &\ 67.4 &\ 67.9 & 67.7 &66.5 &\ 67.2 &\ 67.9 \\
\hline
\end{tabular}
\end{center}
\caption{Ablation study of hyper-parameter settings and choices for producing the box embedding features in the proposed approach. }
\vspace{-1.5em}
\label{table.parameter_abalation}
\end{table*}

To examine the impact of the key components in our integrated object detection and tracking, we perform ablations in an online setting. The results are shown in Table~\ref{table.component_abalation}. The baseline method excludes tracklet-conditioned detection in Algorithm~\ref{alg.integrated_detect_track}, which is equivalent to a basic version of sequential detection and tracking where box propagation and rescoring are removed in Algorithm~\ref{alg.sequential_detect_track}. This baseline obtains mAP$^\text{det}$ and mAP$^\text{track}$ scores of 74.6\% and 65.2\%, respectively. Applying tracklet-conditioned object detection on either the first or second stages of the object detector leads to improvements in mAP$^\text{det}$ and mAP$^\text{track}$ of 0.8\% and 1.9\%, and 2.2\% and 1.6\%, respectively. With tracklet-conditioned detection on both stages, mAP$^\text{det}$ and mAP$^\text{track}$ become 78.1\% and 67.9\%, respectively, which are improvements of 3.5\% and 2.7\% over the baseline.

In the sequential counterpart with late integration, applying box propagation improves the mAP$^\text{det}$ and mAP$^\text{track}$ scores to 75.2\% and 65.9\%, respectively.  Additionally applying online box rescoring as a postprocess improves the mAP$^\text{det}$ score to 75.8\%. The mAP$^\text{track}$ score remains the same, because tracklets are not changed by box rescoring.  To sum up, our full version of integrated detection and tracking outperforms that of sequential detection and tracking with late integration by 2.3\% and 2.0\% in mAP$^\text{det}$ and mAP$^\text{track}$, respectively. 

For a more detailed look at our algorithm's performance, following~\cite{zhu2017flow}, we break down the results into different motion speeds, based on whether the ground-truth tracklet is slow (the mean IoU overlap between boxes in consecutive frames is more than 0.8), medium (0.6$\le$mean IoU$\le$0.8), or fast (mean IoU$<$0.6). As shown in Table~\ref{table.component_abalation}, the gain in mAP$^\text{track}$ over the sequential baseline and late integration grows larger with medium and faster object motion. For these more challenging cases, object detection benefits even more from tracklet information, which in turn leads to improved tracking performance.

We additionally ablate choices for producing the box embedding features used in determining the association weights. The results, displayed in Table~\ref{table.parameter_abalation}, show that adding a separate 2fc head to produce the first stage embedding features leads to better performance. Table~\ref{table.parameter_abalation} also shows ablations over hyper-parameter values. The performance was found to be relatively stable with respect to these values, and the best combination was chosen as the default setting.

\subsection{Tracklet Stability}

We also analyze the tracklet stability of different approaches. In~\cite{zhang2016stability}, the stability of detection boxes in videos was first studied, using proposed metrics that account for temporal stability (fragment error) and spatial stability (box center and aspect ratio errors). Here, we employ slight modifications of these metrics that instead measure tracklet stability. Details on these metrics are given in the Appendix.

Table~\ref{table.stability} compares the difference in stability of our approach to those of sequential detection and tracking with late integration. Our approach reduces the fragment, center and aspect ratio errors by a relative 4\%, 6\% and 7\%, respectively. Improvements in stability are found to be more obvious for objects with fast motion. These numerical results verify the discussion in Section~\ref{sec:discussion}.

\setlength{\tabcolsep}{2pt}
\renewcommand{\arraystretch}{1.0}
\begin{table}[t]
\small
\begin{center}
\begin{tabular}{c|ccc}
\hline
\tabincell{c}{\scriptsize motion split} & \tabincell{c}{\scriptsize frag ($\times 10^{-3}$)} & \scriptsize center ($\times 10^{-3}$) & \scriptsize aspect ($\times 10^{-3}$)\\
\hline
all & \scriptsize $26\xrightarrow{-4\%} 25$ & \scriptsize $134\xrightarrow{-6\%}126$ & \scriptsize  $236\xrightarrow{-7\%}219$ \\
slow & \scriptsize $11\xrightarrow{+27\%}14$ & \scriptsize $88 \xrightarrow{-1\%}87$ & \scriptsize $184\xrightarrow{-8\%}170$ \\
median & \scriptsize $37\xrightarrow{+11\%} 41$ & \scriptsize $173\xrightarrow{-3\%}168$  & \scriptsize $304\xrightarrow{-7\%}282$\\
fast & \scriptsize$63\xrightarrow{-24\%}48$ & \scriptsize$227\xrightarrow{-10\%}205$ & \scriptsize$336\xrightarrow{-6\%}317$ \\
\hline
\end{tabular}
\end{center}
\caption{Tracking stability change from ``sequential detection and tracking with late integration" to ``integrated detection and tracking". `Frag', `center', and `aspect' denote fragment, box center, and aspect ratio errors, respectively. The error numbers are shown in the format of ``sequential with late integration$\xrightarrow{\text{relative change}}$integrated".}
\vspace{-1.0em}
\label{table.stability}
\end{table}

\subsection{Results on Stronger Baselines and Comparison to State-of-the-art Approaches}

In Table~\ref{table.strong_baseline}, our approach is compared to sequential detection and tracking with late integration on stronger baselines. The network features are enhanced by applying combinations of FGFA~\cite{zhu2017flow} and Deformable ConvNets v2~\cite{zhu2018deformable}. On these high baselines, the integrated detection and tracking approach still outperforms the sequential counterpart by a clear margin in both detection and tracking.

\setlength{\tabcolsep}{2pt}
\renewcommand{\arraystretch}{1.1}
\begin{table}[t]
\small
\begin{center}
\begin{tabular}{cc|cc}
\hline
\scriptsize DCNv2 & \scriptsize FGFA & \tabincell{c}{mAP$^{\text{det}}$} & mAP$^{\text{track}}$\\
\hline
& &$75.8 \rightarrow 78.1$ & $65.9 \rightarrow 67.9 $ \\
$\checkmark$ & & $79.4 \rightarrow 82.0$ & $68.4 \rightarrow 70.8$\\
$\checkmark$ & $\checkmark$ & $81.5 \rightarrow 83.5$ & $70.2 \rightarrow 72.6$\\
\hline
\end{tabular}
\end{center}
\caption{Improvement on stronger baselines with  FGFA~\cite{zhu2017flow}, and Deformable ConvNets v2 (DCNv2)~\cite{zhu2018deformable}. The scores are reported in the format of  ``sequential with late integration$\rightarrow$integrated".}
\vspace{-1em}
\label{table.strong_baseline}
\end{table}

\setlength{\tabcolsep}{2pt}
\renewcommand{\arraystretch}{1.0}
\begin{table}[t]
\small
\begin{center}
\begin{tabular}{c|c|c|cc}
\hline
\tabincell{c}{method} & \tabincell{c}{inference} &\tabincell{c}{backbone} & \tabincell{c}{mAP$^{\text{det}}$} &
mAP$^{\text{track}}$\\
\hline
NUIST~\cite{nuist2016vid} & off-line & ensemble            & 81.2 & N.A.   \\
NUS-Qihoo-UIUC~\cite{nus2017vid} & off-line & DPN-131~\cite{chen2017dual}      & 83.1 & 70.3   \\
\hline
FGFA~\cite{zhu2017flow} & off-line & ResNet-101 & 78.4 & N.A. \\
\hline
THP~\cite{zhu2018towards} & on-line & ResNet-101 & 78.6 & N.A. \\
\hline
\multirow{2}{*}{D\&T~\cite{feichtenhofer2017detect}} & on-line & \multirow{2}{*}{ResNet-101} & 78.7 & -   \\
 & off-line &  & 79.8 & -   \\
\hline
D\&T (reproduced) & off-line & ResNet-101 & 79.0 & 60.5   \\
\hline
Ours & on-line & ResNet-101             & \textbf{83.5} & \textbf{72.6}   \\
\hline
\end{tabular}
\end{center}
\caption{Comparison to state-of-the-art systems on the ImageNet VID validation set. In the paper of D\&T~\cite{feichtenhofer2017detect}, the mAP$^{\text{track}}$ score is not reported, so we reproduced the approach and report the results.}
\vspace{-1em}
\label{table.VID_system}
\end{table}

\setlength{\tabcolsep}{1pt}
\renewcommand{\arraystretch}{1.0}
\begin{table}[t]
\small
\begin{center}
\begin{tabular}{c|c|c|cccc}
\hline
\tabincell{c}{method} & \tabincell{c}{inference} &\tabincell{c}{pre-trained} & \tabincell{c}{MOTA} & FP & FN & IDSw\\
\hline
H1\_SJTUZTE~\cite{2dmot15leaderboard} & off-line & unknown & \textbf{56.6} & 7198 & 18926 &  533 \\
RAR15~\cite{fang2018recurrent} & on-line & D+T& 56.5 & 9386 & \textbf{16921} & 428 \\
TRID~\cite{manen2017pathtrack} & off-line & D+T& 55.7 & 6273 & 20611 &  \textbf{351} \\
NOMTwSDP~\cite{choi2015near} & off-line & unknown & 55.5 & 5594 & 21322 &  427 \\
AP\_HWDPL~\cite{chen2017online} & on-line & D+T & 53.0 & \textbf{5159} & 22984 &  708 \\
CDA\_DDAL~\cite{bae2018confidence} & on-line & D+T & 51.3 & 7110 & 22271 &  544 \\
MDP\_SubCNN~\cite{sadeghian2017tracking} & on-line & D& 47.5 & 8632 & 22969 &  628 \\
DMT~\cite{kim2016cdt} & off-line & D& 44.5 & 8088  & 25336 & 684 \\
\hline
Ours & on-line & D & 56.1 & 5717 & 20460 & 788 \\
\hline
\end{tabular}
\end{center}
\caption{Comparison to state-of-the-art systems on the 2D MOT15 test set. `D' and `T' indicate pre-training for the object detection task and the tracking task, respectively.}
\vspace{-1.5em}
\label{table.MOT_system}
\end{table}

We further compare the proposed approach implemented on the highest baseline to the state-of-the-art methods at the system level. Table~\ref{table.VID_system} and Table~\ref{table.MOT_system} present the results. We note that due to system complexity and missing implementation details, direct and fair comparison among different works is difficult. Our system of ``integrated detection and tracking" achieves accuracy that is very competitive with all the other systems. And we note that the idea of early integration and tracklet-conditioned detection should be applicable to these other detection and tracking systems as well.

\section{Conclusion}

Both object detection and tracking are fundamental tasks in video understanding that are closely coupled by nature. However, in the previous approaches, the object detection and tracking modules are applied in a sequential manner, and are optionally integrated at a late stage. In this paper, we propose the first approach to tightly integrate the tasks by conditioning object detection on the current frame by tracklets from the previous frames. The object detection results are not only more accurate, but also more coherent with the existing tracklets, which further improves tracking results. Extensive experiments on the ImageNet VID and the 2D MOT 2015 benchmarks demonstrate the effectiveness of the proposed approach. The idea of early integration and tracklet-conditioned detection can also be applied to other video understanding tasks which involve both recognition and temporal association, such as jointly estimating and tracking human pose.

\appendix
\renewcommand{\thesection}{A\arabic{section}}   
\section{Experimental Setting Details}

\paragraph{ImageNet VID dataset~\cite{ILSVRC15}}  This dataset is a commonly used large-scale benchmark for video object detection and tracking. The training, validation, and test sets contain 3862, 555, and 937 video snippets, respectively. The frame rate is 25 or 30 fps for most snippets. All the object instances are fully annotated with bounding boxes and instance IDs, providing a good benchmark for joint object detection and tracking. There are 30 object categories, which are a subset of the categories in the ImageNet DET dataset. 

Following the protocol in \cite{lee2016multi,zhu2017deep,zhu2017flow}, in all our experiments, the models are trained on the union of the ImageNet VID training set and the ImageNet DET training set (only the same 30 category labels are used), and are evaluated on the ImageNet VID validation set. In both training and inference, the input images are resized to a shorter side of 600 pixels. In RPN, the anchors are of 3 aspect ratios \{1:2, 1:1, 2:1\} and 4 scales \{$64^2$, $128^2$, $256^2$, $512^2$\}. 300 region proposals are generated for each frame at an NMS threshold of 0.7 IoU. SGD training is performed, with one image at each mini-batch. 120k iterations are performed on 4 GPUs, with each GPU holding one mini-batch. The learning rates are $10^{-3}$ and $10^{-4}$ in the first 80k and last 40k iterations, respectively. In each mini-batch, images are sampled from ImageNet DET and ImageNet VID at a 1:1 ratio. The weight decay and the momentum parameters are set to $0.0001$ and $0.9$, respectively. In inference, detection boxes are generated at an NMS threshold of 0.3 IoU.

\paragraph{2D MOT 2015~\cite{leal2015motchallenge}} This dataset is a widely used benchmark for multiple object tracking. It contains a total of 22 videos collected under varying scenes, devices and angles. Only the pedestrians are annotated. These videos are divided into 11 training videos and 11 test videos. The training videos have $5500$ frames, $500$ tracklets, and $39905$ boxes. The test videos have $5783$ frames, $721$ tracklets, and $61440$ boxes. The average number of boxes for each frame is $7.3$ and $10.6$ in the training and test set, respectively. The frame rates of this dataset varies greatly, ranging from $2.5$ fps to $30$ fps. This dataset is very challenging for pedestrian detection and tracking, due to occlusions, high annotation density, high diversity of scenarios, etc. 

In both training and inference, the input images are resized to a shorter side of 800 pixels. Anchors of 3 aspect ratios \{1:2, 1:1, 2:1\} and 5 scales \{$32^2$, $64^2$, $128^2$, $256^2$, $512^2$\} are utilized in RPN. 512 and 2000 region proposals are generated on each frame during training and inference at an NMS threshold of 0.7, respectively. In SGD training on COCO for object detection, 120k iterations are performed on 8 GPUs with 2 images per GPU. The learning rate is initialized to 0.02 and is divided by 10 at the 75k and 100k iterations. In finetuning on 2D MOT 2015 for integrated detection and tracking, 110k iterations are performed on 4 GPUs, with each GPU holding one image. The learning rates are $10^{-3}$ and $10^{-4}$ in the first 70k and last 40k iterations, respectively. The weight decay and the momentum parameters are set to $0.0001$ and $0.9$, respectively. In inference, detection boxes are generated at an NMS threshold of 0.5 IoU. We also utilize common practices developed in previous works~\cite{yu2016poi,xiang2015learning} to better fit the MOTA metric: (1) To reduce FP error, detection boxes with confidence score less than 0.95 are removed, and tracklets with length less than 10 frames are removed; (2) To reduce IDSw error, in online processing, previous tracklets not associated with boxes for 10 consecutive frames are not allowed to be associated with any new boxes in the upcoming frames (but the tracklets are kept in the final results).

\section{Tracklet Stability Metric}

In~\cite{zhang2016stability}, the authors first examined the problem of detection and tracking stability. Three metrics are proposed for evaluating stability, namely, fragment error, center position error, and scale and aspect ratio error. The metrics are applied on the per-frame detection boxes, produced by video object detection algorithms. For stability evaluation, the detection boxes are assigned to pseudo tracklets, aided by the oracle ground-truth annotations. For each ground-truth tracklet, a pseudo tracklet is formed approximately by picking the detection box with the highest overlap with respect to the corresponding ground-truth at each frame\footnote{The implementation in~\cite{zhang2016stability} performs maximum bipartite graph matching with the box IoUs as the weights of the bipartite graph.}. The stability errors are averaged over all the pseudo tracklets. It is not specified in~\cite{zhang2016stability} how to extend their approach to tracklets produced by detection and tracking algorithms.

Here, we extend~\cite{zhang2016stability} for evaluating the stability of detection and tracking algorithms in a straightforward way. Similar to the approach in~\cite{zhang2016stability}, we seek to find a ``best-match" tracklet for each ground-truth tracklet. All the recognized tracklets are first classified into positive and negative tracklets, according to the box IoU and temporal IoU thresholds in the mAP$^{\text{track}}$ metric. A positive tracklet is assigned to the ground-truth tracklet with the highest temporal IoU. For each ground-truth tracklet, the tracklet with the highest classification score among all its assigned tracklets is picked as its ``best-match". The resulting stability errors are the averaged errors over all the ``best-match" tracklets (generated at various box and temporal IoU thresholds as done for mAP$^{\text{track}}$).

{\small
\bibliographystyle{ieee}
\bibliography{egbib}
}

\end{document}